\documentclass{jmlr}
\makeatletter
\providecommand{\@jmlr@authors@sep}{, }
\makeatother

\usepackage{graphicx}
\usepackage{subcaption}
\usepackage{amssymb}
\usepackage{longtable}
\usepackage{bbm}

\usepackage{booktabs}
\usepackage{lineno}

\makeatletter
\let \@jmlrpages \@empty
\let\Ginclude@graphics\@org@Ginclude@graphics 
\makeatother

\jmlrvolume{304,}
\jmlryear{2025}
\jmlrworkshop{ACML 2025}

\title[Balanced Knowledge Updates in LLMs]{Balancing Knowledge Updates: Toward Unified Modular Editing in LLMs}

\author{
  \Name{Jiahao Liu} \Email{ljiahao@stu2023.jnu.edu.cn}\\
  \addr School of Intelligent Systems Science and Engineering, Jinan University
  \AND
  \Name{Zijian Wang} \Email{zwan0998@uni.sydney.edu.au}\\
  \addr School of Computer Science, The University of Sydney
  \AND
  \Name{Kuo Zhao}\footnotemark[1] \Email{zhaokuo@jnu.edu.cn}\\
  \Name{Dong Hu} \Email{hudong@stu2023.jnu.edu.cn}\\
  \addr School of Intelligent Systems Science and Engineering, Jinan University
}

\editors{Hung-yi Lee and Tongliang Liu}

\begin{document}
\begingroup
\renewcommand\thefootnote{\fnsymbol{footnote}} 
\maketitle
\makeatletter
\let \@jmlrpages \@empty
\makeatother
\footnotetext[1]{Corresponding author}
\endgroup

\begin{abstract}
Knowledge editing has emerged as an efficient approach for updating factual knowledge in large language models (LLMs), typically achieved by first locating key knowledge-storage modules and then modifying their parameters. 
However, most existing methods focus exclusively on updating the weights of Multi-Layer Perceptron (MLP) modules, which are commonly identified as the primary repositories of factual information.
Other important components, such as attention (Attn) modules—one of the core modules in LLMs—are often ignored during editing. 
This biased allocation of updates can leave residual outdated knowledge in the model and limit the effectiveness of knowledge editing.
In this paper, we conduct comprehensive and systematic knowledge localization experiments on advanced LLMs, revealing that Attn modules play a substantial role in factual knowledge storage and retrieval, especially in earlier layers. 
Building on these insights, we propose \textit{IntAttn-Edit}, a novel method that extends the associative memory paradigm to jointly update both MLP and Attn modules. 
Our approach employs a knowledge balancing strategy that proportionally allocates update magnitudes based on each module’s measured contribution to knowledge storage. 
Extensive experiments on popular benchmarks demonstrate that \textit{IntAttn-Edit} consistently achieves superior results over existing methods, delivering higher edit success, improved generalization, and robust knowledge preservation.
Further empirical analysis shows that our knowledge balancing strategy enables the editing performance to remain within the optimal range across different settings.
\end{abstract}

\section{Introduction}
Large language models (LLMs) have made remarkable progress in natural language understanding and generation. 
However, despite these advances, LLMs remain susceptible to factual inaccuracies, can generate potentially harmful or unsafe content, and often encode outdated knowledge due to their fixed training cut-off~\cite{zhang2023large,chen2024combating}. 
Retraining LLMs to address these limitations is both expensive and time-consuming~\cite{zheng2023learn,liu2021towards}. 
As a result, there has been a growing interest in knowledge editing techniques, which aim to enable precise, efficient, and cost-effective post-hoc updates to factual knowledge~\cite{yao2023editing,wang2024knowledge}. Such methods allow for the targeted refinement of specific model behaviors without extensive retraining or sacrificing overall performance.

Most mainstream knowledge editing methods adopt a locate-then-edit paradigm~\cite{meng2022locating,meng2022mass}: they first identify the specific modules and layers that store the target factual knowledge, and then use these localization results to inform where to modify the parameters for knowledge updates.
However, the dominant informing strategy is winner-takes-all: only the most significant knowledge storage modules (“winners”) are updated, while other knowledge-bearing modules (“losers”) remain unchanged.
Specifically, extensive studies~\cite{geva2020transformer, geva2022transformer} have shown that factual knowledge in large language models is primarily stored in MLP modules. The most popular editing methods, such as ROME~\cite{meng2022locating}, MEMIT~\cite{meng2022mass}, and AlphaEdit~\cite{fang2024alphaedit}, focus exclusively on updating MLP parameters.
However, this biased winner-takes-all strategy neglects other important modules—such as Attn—which may result in residual outdated knowledge and prevent LLMs from fully acquiring new information. Recently, there has been a surge of research focusing on attention mechanisms~\cite{vaswani2017attention, shazeer2019fast, ainslie2023gqa, liu2024deepseek}, with mounting evidence showing that Attn layers are also crucial for encoding and retrieving knowledge~\cite{hase2023does}, especially in advanced LLMs. These findings collectively indicate that attention, as a fundamental component of Transformer architectures, should not be ignored in knowledge editing.

In this paper, we first conduct a systematic and comprehensive investigation of knowledge localization in advanced LLMs using causal tracing. 
Our experiments demonstrate that not only do MLP modules serve as major repositories of factual knowledge, but Attn modules also play an equally critical role in knowledge storage. 
Motivated by these findings, we propose \textit{IntAttn-Edit} (Integrating Attention to Edit), a novel method that, for the first time, simultaneously updates the parameters of both MLP and Attn modules during knowledge editing.
Unlike previous approaches that exclusively update the dominant (“winner”) module, \textit{IntAttn-Edit} employs a knowledge balancing strategy that dynamically allocates update magnitudes between the two modules in proportion to their empirically measured contributions to factual knowledge. 
This dual-module strategy ensures that all relevant knowledge storage components are effectively updated, thereby reducing knowledge residuals and consistently yielding superior editing performance compared to single-module methods.

In summary, our main contributions are as follows:
\begin{itemize}
    \item[(1)]  Through extensive and comprehensive knowledge localization experiments, we discover that Attn modules also play a vital role in factual knowledge encoding. 
    Specifically, Attn modules can function as key-value associative memory structures, dynamically encoding factual associations alongside MLP modules.
    \item[(2)] We propose \textit{IntAttn-Edit}, a novel approach that extends the key-value memory editing paradigm from MLP layers to Attn modules for the first time. 
    By introducing a knowledge balancing strategy, \textit{IntAttn-Edit} dynamically allocates update magnitudes between MLP and Attn modules based on their empirical contributions to knowledge storage. 
    This ensures that all major knowledge-bearing components are effectively updated, reducing knowledge residuals and improving the overall robustness of knowledge editing.
    \item[(3)] Through extensive experiments on advanced LLMs and popular benchmarks, we demonstrate that \textit{IntAttn-Edit} consistently surpasses leading editing baselines across key performance metrics, particularly in batch editing scenarios. 
    Furthermore, ablation studies confirm that our knowledge balancing strategy maintains editing effectiveness within an optimal range by dynamically allocating update magnitudes according to each module’s empirical contribution, underscoring the robustness and practical value of our approach.
\end{itemize}

\section{Related Works}

Knowledge editing in large language models (LLMs) aims to efficiently update internal knowledge while maintaining overall model performance. Existing approaches are generally divided into three categories: (1) External Knowledge Editing, (2) Knowledge Integration, and (3) Intrinsic Editing.

\subsection{External Knowledge Editing}

LLMs possess in-context learning abilities, allowing for external knowledge integration through memory mechanisms or model augmentation. 
Representative methods include IKE~\cite{zheng2023can}, which employs diverse demonstrations to guide factual updates, and MeLLo~\cite{zhong2023mquake}, which decomposes multi-hop queries and retrieves answers from sub-question memories. PokeMQA~\cite{Pokemqa} enhances reliability using programmable detectors and knowledge prompts, while SERAC~\cite{mitchell2022memory} combines classifiers with counterfactual models.
However, these methods are limited by retrieval quality, susceptibility to harmful content, and potential knowledge conflicts~\cite{wang2023self, liu2023recall, wang2023resolving}. 
Yu~\cite{yu2023characterizing} further examines the preference for context-based versus memory-based answers. 
In summary, external knowledge-guided techniques support knowledge editing but are constrained by the consistency and accuracy of retrieval.

\subsection{Knowledge Integration}

Knowledge integration approaches introduce new knowledge representations by adding or interpolating components at output or intermediate layers.
Examples include knowledge patches as alternative output heads~\cite{murty2022fixing}, parameter-efficient adapters like LoRA~\cite{hu2022lora}, and dynamic plug-in modules such as MELO~\cite{yu2024melo}. GRACE~\cite{hartvigsen2023aging} utilizes a dynamic codebook for continual adaptation.
Although these methods offer flexible knowledge updates, they must carefully manage knowledge conflicts and model capacity to ensure effective integration from multiple sources.

\subsection{Intrinsic Editing}

Intrinsic editing directly updates model parameters to encode new knowledge internally. 
Although full-model fine-tuning is resource intensive and prone to catastrophic forgetting, regularized strategies like Constrained Finetune~\cite{zhu2020modifying} mitigate such issues. 
Meta-learning approaches (e.g., KE~\cite{de2021editing}, SLAG~\cite{hase2023methods}) and location-then-edit paradigms (e.g., Knowledge Neuron~\cite{dai2021knowledge}, ROME~\cite{meng2022locating}) enable more focused edits. 
Low-rank methods like MEND~\cite{mitchell2021fast} reduce parameter cost, yet challenges remain in handling multi-edit conflicts and preserving non-target knowledge.
Location-then-edit strategies identify critical storage sites for precise updates: Knowledge Neuron attributes knowledge via gradient sensitivity, while ROME leverages causal analysis for minimal-weight optimization. Recent advances such as PMET~\cite{li2024pmet}, R-ROME~\cite{gupta2024rebuildingromeresolving}, and AlphaEdit~\cite{fang2024alphaedit} further enhance editing precision and locality.

\section{Preliminaries}
\subsection{Knowledge Location in Transformer}
\paragraph{Autoregressive Language Models.}
An autoregressive language model $\mathcal{F}_\theta$ predicts the probability distribution of the next token in a sequence by modeling dependencies among preceding tokens.
Given an input sequence $[x_1, x_2, ..., x_z]$, the model generates the next token according to the conditional probability $\mathbb{P}(x_{z+1} \mid x_1, ..., x_z)$, which is parameterized by a $D$-layer Transformer decoder~\cite{vaswani2017attention}:
\begin{equation}
    \mathcal{F}_\theta(x_1, ..., x_z) = \mathrm{softmax}(W_E \gamma(h^D_z)) = \mathbb{P}(x_{z+1}|x_1, ..., x_z),
\end{equation}
where $h^D_z$ denotes the hidden state at position $z$ in the final ($D$-th) layer, $W_E$ is the output embedding matrix, and $\gamma$ represents layer normalization.

The hidden state $h_z^l$ for token position $z$ at layer $l$ is computed recursively as follows:
\begin{equation}
h_z^l = h_z^{l-1} + a_z^l + m_z^l
\end{equation}
where $a_z^l$ and $m_z^l$ denote the outputs of the Attn and MLP modules, respectively. 
The process can be expanded as:
\begin{equation}
a^l_i = W_o^l \cdot \mathrm{ATTN}^l \left( \gamma\left(h_1^{l-1}, h_2^{l-1}, ..., h_i^{l-1}\right) \right)
, \quad
m^l_i = W_{\mathrm{out}}^l \sigma\left( W_{\mathrm{in}}^l \gamma\left(h_i^{l-1}\right) \right)
\end{equation}
where $W_o^l$ is the Attn output projection matrix, $\mathrm{ATTN}^l(\cdot)$ denotes the multi-head self-attention operation at the $l$-th layer, $W_{\mathrm{in}}^l$ and $W_{\mathrm{out}}^l$ are the input and output matrices of the MLP, and $\sigma$ is a non-linear activation function.

\vspace{0.3em}
\paragraph{Causal Tracing for Knowledge Localization.}
Causal tracing~\cite{meng2022locating} provides an effective means for identifying the storage and processing locations of factual knowledge within Transformer models.
This procedure involves constructing a clean prompt ($X_{\mathrm{clean}}$) with its associated answer $r$, a corrupted prompt ($X_{\mathrm{cor}}$), and executing three model runs:

\begin{enumerate}
    \item[(1)] \textbf{Clean run}: Execute the model on $X_{clean}$ to record intermediate activations of specific computational components.
    \item[(2)] \textbf{Corrupted run}: Process $X_{cor}$ to generate predictions under knowledge-disrupted conditions. 
    To generate $X_{cor}$, Gaussian noise $\mathcal{N}(0, \nu)$ is added to the embeddings of some key tokens, where $\nu$ is 3 times the standard deviation of the token embedding text set.
    \item[(3)] \textbf{Patched run}: Intervene by reinstating cached activations from $X_{clean}$ to a designated component during $X_{cor}$ inference, while preserving corruption in other components.
    Output deviations quantify the component's causal role in factual recall.
\end{enumerate}

Prior causal tracing studies~\cite{meng2022locating,meng2022mass} on classical models such as GPT-2 have shown that MLP modules, particularly their output projections, are pivotal for factual memory retrieval. 
As a result, most knowledge editing methods~\cite{fang2024alphaedit,gupta2024rebuildingromeresolving} have focused primarily on identifying and updating MLP output parameters to insert or modify facts. 
In contrast, despite being a central component of Transformer architectures, the Attn module’s role in knowledge storage has remained largely underexplored in mainstream knowledge editing research.

\subsection{Linear associate memory and knowledge update}
\paragraph{Linear associative memory.}
Most editing methods interpret the output weight matrices of FFN layers, $W_{\mathrm{out}}$, as linear associative memories~\cite{geva2020transformer}. 
In this view, any linear operation $W$ serves as a parameterized key–value store, where a matrix of keys $K = [k_1\ |\ k_2\ |\ \ldots ]$ maps to a set of values $V = [v_1\ |\ v_2\ |\ \ldots ]$, aiming for $W K \approx V$. 
To best fit these associations, $W$ is typically solved by minimizing the squared error, resulting in the closed-form solution $W = V K^+$ using the Moore–Penrose pseudoinverse.

Some researches~\cite{bau2020rewriting,geva2020transformer} show that a new key–value pair $(k^*, v^*)$ can be optimally integrated into the associative memory by solving a constrained least-squares problem. Specifically, the updated weight matrix $\hat{W}$ is required to satisfy:
\begin{equation}
\hat{W} k^* = v^*, \quad \text{where} \quad \hat{W} = W + \Delta
\end{equation}

The goal is thus to solve for the update $\Delta$ that enables the direct insertion of $(k^*, v^*)$ while minimally disturbing the original key–value structure.




\paragraph{Model Editing in LLMs.}

Knowledge editing in LLMs updates factual associations by directly modifying the parameters responsible for key-value mappings.
When knowledge is represented as (subject, relation, object) triples (e.g., $s$ = "LeBron James", $r$ = "plays sport", $o$ = "basketball"), the MLP maps semantic input to the factual object. 
In particular, the output weight matrix $W_{\mathrm{out}}^l$ in specific MLP layers is frequently interpreted as a linear associative memory. 
Within this framework, the key vector for each input, $\mathbf{k} = \sigma(W_{\mathrm{in}}^l\gamma(h^{l-1}))$, encodes the relevant (subject, relation) semantics. 
The MLP output $m^l$ serves as the value corresponding to the target knowledge, i.e., $\mathbf{v} = m^l$.
For convenience, we denote $W_0$ as $W_{\text{out}}^l$ in the following.

Suppose each editing operation aims to update $u$ new pieces of knowledge, each in the form of a ($s$, $r$, $o$) triple. 
Given the new knowledge key-value pairs, we seek a parameter update $\Delta$ that encodes the new facts while preserving the irrelevant existing knowledge.
Let $K_1 = [k_1, \dots, k_u] \in \mathbb{R}^{d_k \times u}$ and $V_1 = [v_1, \dots, v_u] \in \mathbb{R}^{d_v \times u}$ denote the key and value matrices for the new knowledge, where $d_k$ and $d_v$ are the dimensions of keys and values, respectively. 
Similarly, $(K_0, V_0)$ represent the key-value pairs corresponding to the original knowledge that should be maintained. 
Current methods~\cite{gu2024model,meng2022mass} typically formulate editing as an optimization problem, seeking a perturbation $\Delta$ to update $W_0$ so that the model’s new associations match the targets while preserving previous knowledge:
\begin{equation}
\Delta = \arg\min_{\tilde{\Delta}} \left(
\left\| (W_0 + \tilde{\Delta}) K_1 - V_1 \right\|^2 +
\left\| (W_0 + \tilde{\Delta}) K_0 - V_0 \right\|^2
\right)
\end{equation}

This admits a closed-form solution:
\begin{equation}
\Delta = (V_1 - W_0 K_1) K_1^{\top} \left( K_0 K_0^{\top} + K_1 K_1^{\top} \right)^{-1}
\label{eq:delta}
\end{equation}

This framework is naturally extendable to multi-layer scenarios. 
In such cases, the residual $R$ (desired minus current outputs) is distributed across critical MLP layers.
Let $\mathcal{S}$ denote these layers, $l \in \mathcal{S}$, and $L \triangleq \max(\mathcal{S})$ the highest index. 
For each layer $l$, the residual and incremental weight update are computed as

\begin{equation}
R^{l} = \frac{V_1 - W_0 K_1}{L - l + 1}, \quad \Delta^{l} = R^{l} K_1^{\top} (K_0 K_0^{\top} + K_1 K_1^{\top})^{-1}
\label{eq:multi_delta}
\end{equation}

Here, $(K_0, V_0)$ represent the key-value pairs of existing knowledge to be preserved, while $(K_1, V_1)$ correspond to the new knowledge being introduced. 
In practice, obtaining the complete $K_0$ that covers all knowledge stored in an LLM is infeasible. 
To approximate $K_0$, previous studies~\cite{meng2022mass} suggest randomly sampling a large number of (subject, relation, object) triplets from external corpora such as Wikipedia. 
For example, it is common to use 100,000 triplets, resulting in $K_0 \in \mathbb{R}^{d_k \times 100{,}000}$.

\section{Methodology}

\subsection{Discovering Attention as Memory in LLMs}

In this section, we revisit the mechanisms underlying factual knowledge localization within LLMs through a systematic causal tracing analysis on a advanced model, Qwen2.5-7B. 
Specifically, we employed activation patching~\cite{meng2022locating} to quantify the causal contributions of both the MLP and Attn modules across different layers and token positions. 

To quantify the causal contribution of model components to factual recall, we employ two core metrics \cite{zhang2023towards} comparing model performance between \textit{corrupted} ($*$) and \textit{patched} ($pt$) runs:

\begin{itemize}
    \item \textbf{Probability}: The patching effect is defined as $P(r) = P_{pt}(r) - P_{*}(r)$, where $r$ denotes the correct answer.
    \item \textbf{Logit difference}: The normalized restoration of the logit gap between the target answer $r$ and a contrastive incorrect answer $r'$, quantifying the component-specific corrective capability:
    \begin{equation}
        LD(r, r') = \frac{LD_{pt}(r, r') - LD_{*}(r, r')}{LD_{cl}(r, r') - LD_{*}(r, r')}
    \end{equation}
    where $LD(r, r') = \textit{Logit}(r) - \textit{Logit}(r')$, and $r'$ is a frequent incorrect answer. Here, \textit{Logit}$(\cdot)$ refers to the unnormalized output score for each candidate answer produced by the model before applying the softmax function.
\end{itemize}

\begin{figure}[htbp]
  \centering
  \begin{minipage}[t]{0.48\textwidth}
    \centering
    \includegraphics[width=\textwidth]{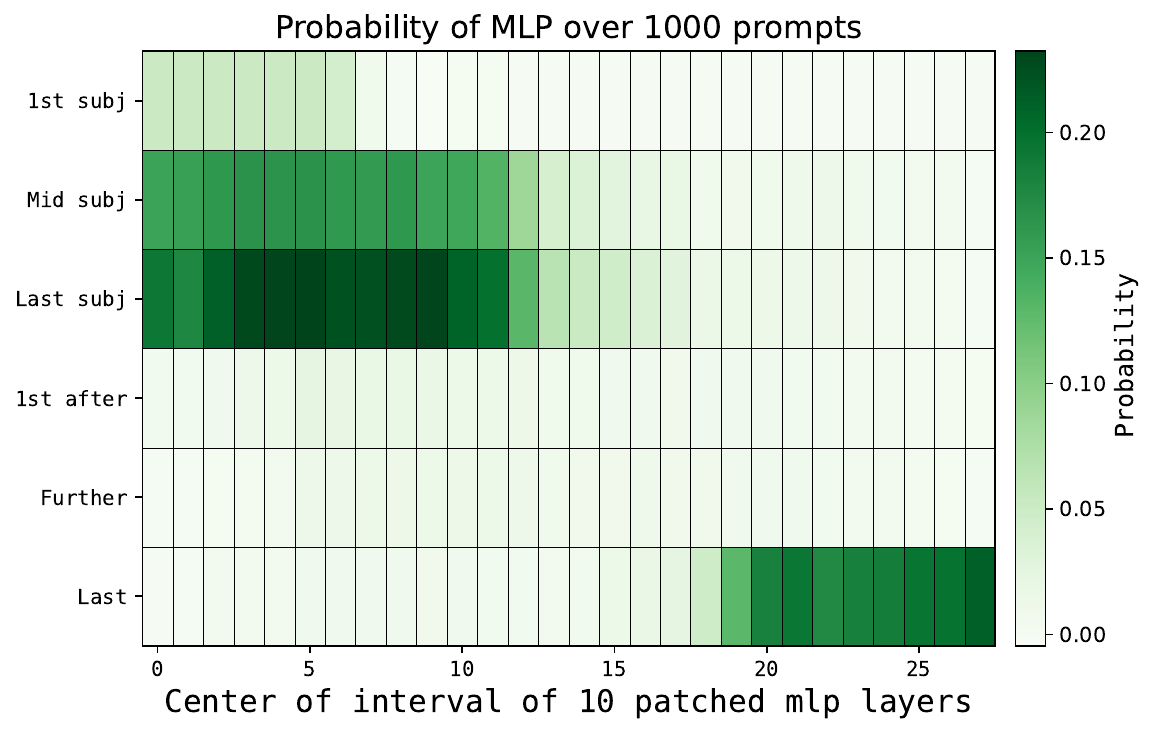}
    \caption*{(a) Probability of MLP over 1000 prompts}
    \label{fig:causal_tracing_results:a}
  \end{minipage}
  \hfill
  \begin{minipage}[t]{0.48\textwidth}
    \centering
    \includegraphics[width=\textwidth]{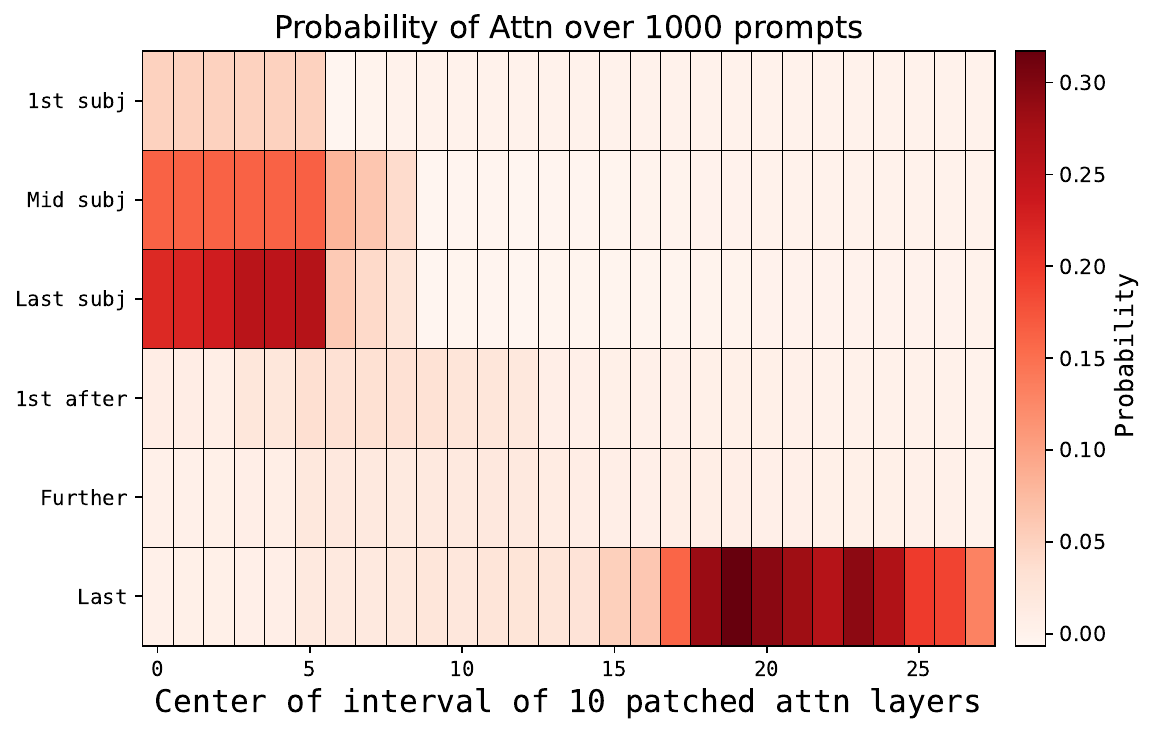}
    \caption*{(b) Probability of Attn over 1000 prompts}
    \label{fig:causal_tracing_results:b}
  \end{minipage}
  \caption{Heatmaps of causal influence from MLP and Attn layers on factual recall in Qwen2.5-7B. }
  \label{fig:causal_tracing_results}
\end{figure}

As shown in Figure~\ref{fig:causal_tracing_results}, the MLP module displays strong causal effects in the middle layers, particularly at the last subject token, which is consistent with previous studies.
In contrast, the Attn module exhibits pronounced causal influence in the early layers (approximately layers 1 to 5), especially at the middle and last subject token positions. 
This highlights the critical role of Attn in establishing relational pathways for factual recall at the initial stages of processing, rather than merely aggregating contextual information.
Together, these results underscore that both MLP and Attn modules play essential roles in the storage and retrieval of factual knowledge within LLMs.

\begin{figure}[htbp]
  \centering
  \begin{minipage}[t]{0.48\textwidth}
    \centering
    \includegraphics[width=\textwidth]{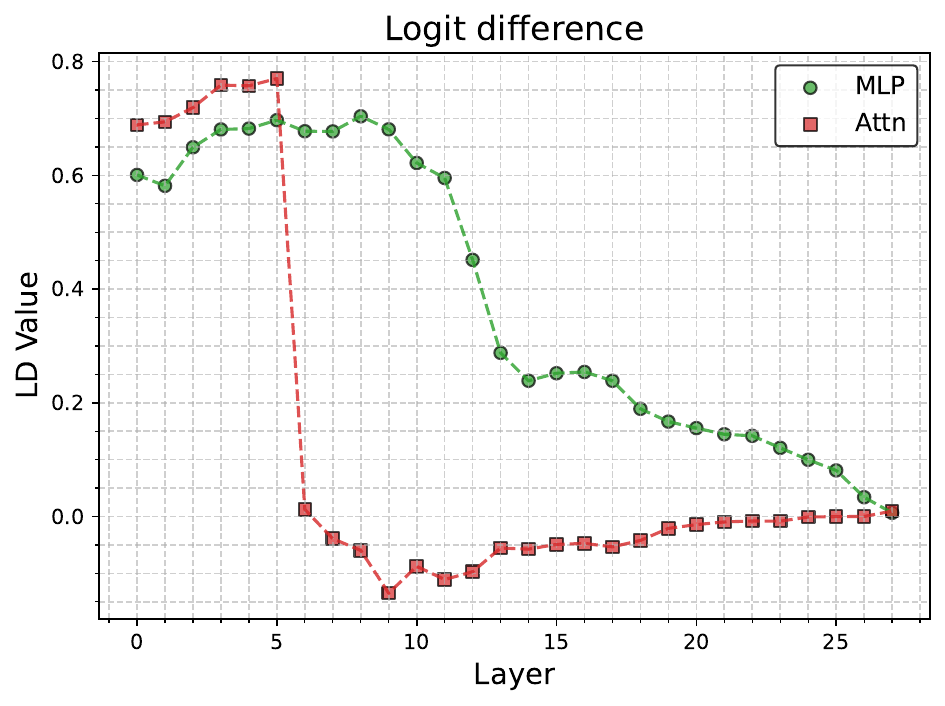}
    \caption*{(a) Logit Difference}
  \end{minipage}
  \hfill
  \begin{minipage}[t]{0.48\textwidth}
    \centering
    \includegraphics[width=\textwidth]{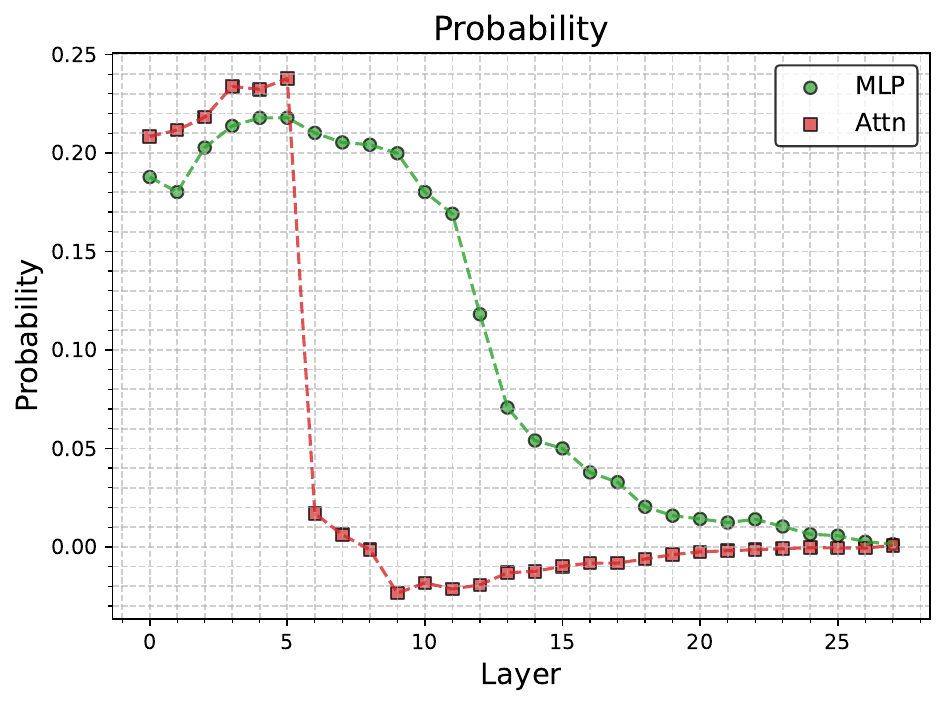}
    \caption*{(b) Probability}
  \end{minipage}
  \caption{Layerwise Causal Effects of Attn and MLP at the Last Subject Token.}
  \label{fig:causal_last_token}
\end{figure}

A closer inspection of Figure~\ref{fig:causal_last_token} reveals that the Attn module exerts its strongest causal influence in the earliest layers, as indicated by a rapid increase and subsequent sharp decline in both logit difference and probability metrics. 
The contribution of Attn to factual recall peaks within the first five layers and then quickly diminishes.
This pattern suggests that \textbf{Attn plays a vital role in the initial identification and encoding of factual associations}, functioning as a semantic filter and router early in the model’s processing pipeline. 
These results highlight the importance of recognizing and leveraging the knowledge storage function of Attn modules—alongside MLP modules—when considering the mechanisms of knowledge representation and update in large language models.

Motivated by these empirical observations, we find that Attn modules also play a crucial role in factual knowledge encoding, particularly in the early layers.
This suggests that Attn modules may function as key-value associative memory structures, dynamically encoding factual associations alongside MLPs. Based on this insight, we explicitly incorporate Attn modules into the linear associative memory framework for knowledge editing. 
To distinguish between layer indices, we denote critical Attn layers as $l_\mathrm{attn}$ and key MLP layers as $l_\mathrm{mlp}$, which are independently determined via causal tracing and generally correspond to different sets of layers within the transformer architecture.

\begin{equation}
    \underbrace{a_{i}^{l_\mathrm{attn}}}_{v} =
    W^{l_\mathrm{attn}} \cdot
    \underbrace{
        \mathrm{ATNN}^{l_\mathrm{attn}}
        \left(
            \gamma(h_{1}^{l_\mathrm{attn}-1},\, h_{2}^{l_\mathrm{attn}-1},\, \ldots,\, h_{i}^{l_\mathrm{attn}-1})
        \right)
    }_{k}
\end{equation}

Here, the hidden states input to the Attn module constitute the key ($k$), and the module’s output is linearly projected by the weight matrix $W^{l_\mathrm{attn}}$ to yield the corresponding value ($v$).
For convenience, we denote $W^{l_\mathrm{attn}}$ as $W_{\text{o}}^l$ in the following.

\subsection{IntAttn-Edit}

\begin{figure}[htbp]
\centering
\includegraphics[width=0.9\linewidth]{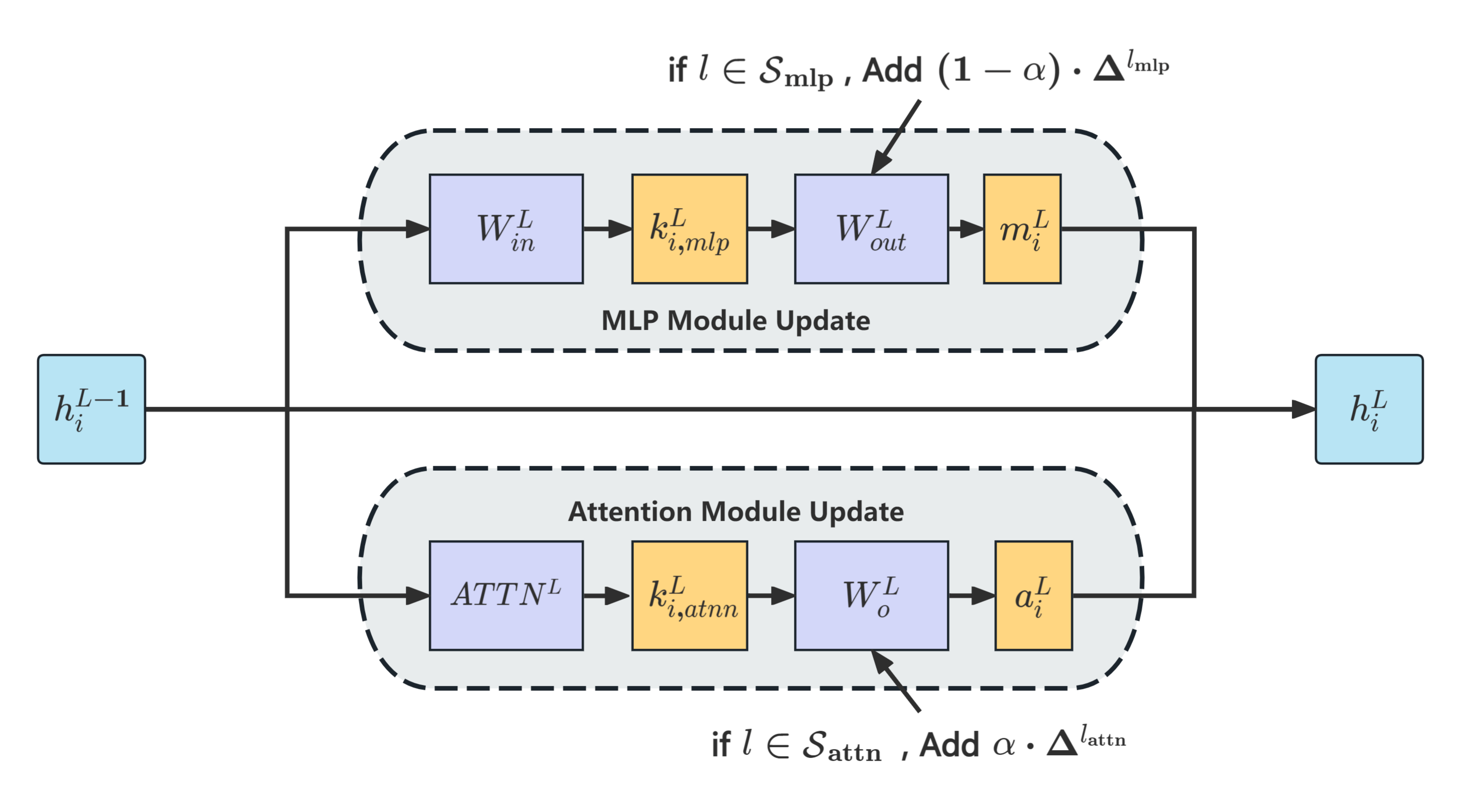}
\caption{Schematic illustration of the dual-pathway parameter update mechanism in \textit{IntAttn-Edit}. 
Both MLP and Attn modules are edited in parallel, with update strengths modulated by the allocation factor $\alpha$ based on the measured causal contributions of each module.}
\label{fig:intattn_edit}
\end{figure}

In this section,\textbf{ we present \textit{IntAttn-Edit}, a novel model editing approach that, for the first time, explicitly targets both the Attn and MLP modules for parameter updates}. 
Unlike prior methods that focus solely on the MLP layers, \textit{IntAttn-Edit} simultaneously performs associative memory updates within both modules. Drawing on empirical evidence that Attn modules can serve as associative memories, we dynamically construct attention-specific key-value pairs to facilitate targeted edits. 
Crucially, to achieve balanced and effective knowledge modification across both pathways, \textbf{we introduce a knowledge balancing strategy that proportionally allocates update magnitudes based on the cumulative causal contributions of each module.} 
As illustrated in Figure~\ref{fig:intattn_edit}, this dual-pathway mechanism ensures that parameter updates are adaptively scaled according to the empirical importance of the MLP and Attn modules, enabling comprehensive and robust knowledge editing.

\paragraph{Deriving Attn Knowledge K-V Pairs.}

For each critical Attn layer $l_{\mathrm{attn}}$, we construct key and value pairs analogously to the MLP case. The keys are derived from the positionally encoded hidden states of the previous layer, while the values are obtained via the Attn output projection. 
Specifically,
\begin{equation}
k_{\mathrm{attn}} = \mathrm{ATTN}^{l_{\mathrm{attn}}} \left( \gamma \left(h_1^{l_{\mathrm{attn}}-1}, \dots, h_i^{l_{\mathrm{attn}}-1}\right) \right), \quad v_{\mathrm{attn}} = W^{l_{\mathrm{attn}}} k_{\mathrm{attn}}
\end{equation}
where $h_i^{l_{\mathrm{attn}}-1}$ denotes the hidden states and $W^{l_{\mathrm{attn}}}$ is the Attn projection matrix.

Once the key-value matrices $K_{1,\mathrm{attn}}$ and $V_{1,\mathrm{attn}}$ for new knowledge are constructed, we compute the update $\Delta^{l_{\mathrm{attn}}}$ for a single Attn layer by solving the same form of closed-form equation as previously established (see Eq.~\eqref{eq:delta}). 
For multi-layer scenarios, this update is broadcast across the designated critical Attn layers following the proportional residual allocation strategy (see Eq.~\eqref{eq:multi_delta}).

Finally, the updated Attn projection weights for layer $l_{\mathrm{attn}}$ are given by:
\begin{equation}
\hat{W}^{l_{\mathrm{attn}}} = W^{l_{\mathrm{attn}}} + \Delta^{l_{\mathrm{attn}}}
\end{equation}

By formulating the Attn module within the linear associative memory paradigm, we enable direct knowledge editing for Attn layers through explicit key-value association updates.

\paragraph{Knowledge Balancing Strategy.}
In this section, \textbf{we propose a Knowledge Balancing Strategy to incorporate all knowledge-storage modules—both Attn and MLP—into the knowledge editing process. }
This strategy ensures that the parameter updates for each module are dynamically allocated in proportion to their empirically measured contributions to factual knowledge. 
It directly addresses a fundamental limitation of prior methods, which typically focus only on updating the dominant module while neglecting the potentially significant role of other modules.

A core challenge is how to quantitatively assess and balance the contributions of different modules to factual recall during editing. 
To solve this, we leverage causal tracing to measure the actual impact (i.e., causal effect) of each module on factual prediction.
Specifically, for each module (MLP or Attn), we compute the cumulative logit difference restoration, $LD^{(l)}(r, r')$, across all layers of that module. 
This metric quantifies how much restoring the activations of a given layer contributes to recovering the correct factual output, thereby providing a direct empirical estimate of the module's knowledge storage responsibility.

Based on these causal effect measurements,\textbf{ we define a balance factor $\alpha$, which represents the proportion of total knowledge storage attributable to the Attn module relative to the combined storage of both modules.} 
The calculation is as follows:
\begin{equation}
\alpha = \frac{\sum_{l \in \mathrm{Attn}} LD^{(l)}(r, r')}
{\sum_{l \in \mathrm{MLP}} LD^{(l)}(r, r') + \sum_{l \in \mathrm{Attn}} LD^{(l)}(r, r')}
\end{equation}
where $LD^{(l)}(r, r')$ denotes the normalized restoration of the logit gap for each layer $l$, quantifying its specific causal contribution to factual recall.

This causal-effect-guided allocation offers several advantages.
First, it allows our method to adaptively allocate update magnitudes between Attn and MLP modules according to their real knowledge contribution, rather than following a fixed or arbitrary rule. 
Second, by distributing updates in proportion to each module's role in knowledge storage, our approach reduces the risk of knowledge residuals—that is, leftover or conflicting information in modules that were ignored by prior winner-takes-all strategies. 
Finally, this fine-grained balancing ensures that all major knowledge-storage components are effectively synchronized during editing, resulting in superior factual accuracy, model consistency, and generalization in practice.

The final parameter updates are then scaled according to this balance factor, as detailed in the following equations:
\begin{equation}
\hat{W}^{l_{\mathrm{mlp}}} = W^{l_{\mathrm{mlp}}} + (1-\alpha)\Delta^{l_{\mathrm{mlp}}}, \quad
\hat{W}^{l_{\mathrm{attn}}} = W^{l_{\mathrm{attn}}} + \alpha\Delta^{l_{\mathrm{attn}}}
\end{equation}
Here, $W^{l_{\mathrm{mlp}}}$ and $W^{l_{\mathrm{attn}}}$ denote the output weight matrices of the MLP and Attn layers, respectively.

\section{Experiments}
\subsection{Exprimental Setup}

\noindent \textbf{Base LLMs and Baseline Methods.}
We choose trending autoregressive LLM models \textbf{Mistral-7B}~\cite{jiang2023mistral7b}, and \textbf{Qwen2.5-7B}~\cite{qwen2025qwen25technicalreport} for evaluation.
We compare our method against several model editing baselines, including \textbf{Fine-Tuning (FT)}~\cite{zhu2020modifying}, \textbf{ROME}~\cite{meng2022locating}, \textbf{R-ROME}~\cite{gupta2024rebuildingromeresolving}, \textbf{MEMIT}~\cite{meng2022mass}, and \textbf{AlphaEdit}~\cite{fang2024alphaedit}. 
All methods are implemented using EasyEdit\cite{zhang2024comprehensive}.

\noindent \textbf{Datasets and Metrics.}
We evaluate our method using two widely adopted benchmarks: the \textbf{ZsRE} dataset~\cite{levy2017zero} and the \textbf{WikiData counterfact} dataset~\cite{meng2022locating}.
We employ \textbf{Edit Success}, \textbf{Portability}, \textbf{Locality}, and \textbf{Fluency} as evaluation metrics~\cite{zhang2024comprehensive}.
The formal definitions are as follows:

\begin{itemize}
    \item \textbf{Edit Success} (\textbf{Edit Succ.}) measures whether the edited model outputs the correct answer for the edited knowledge:
    \begin{equation}
        \text{Edit Succ.} = \frac{1}{N} \sum_{(x_k, y_k^*)} 
        \mathbbm{1}\left\{ \arg\max_{y} f_{\theta'}(y \mid x_k) = y_k^* \right\}
    \end{equation}
    where $x_k$ and $y_k^*$ are the $k$-th edit prompt and its ground-truth answer, $f_{\theta'}$ is the post-edit model, and $N$ is the number of edited samples.

    \item \textbf{Portability} evaluates the accuracy of the post-edited model on the portability set:
    \begin{equation}
        \text{Portability} = \frac{1}{N} \sum_{(x_k, y_k^*)} 
        \mathbbm{1}\left\{ \arg\max_{y} f_{\theta'}(y \mid x_k) = y_k^* \right\}
    \end{equation}
    where $(x_k, y_k^*)$ now refer to portability prompts and their gold answers.

    \item \textbf{Locality} measures the extent to which the model preserves its original predictions on unrelated samples:
    \begin{equation}
        \text{Locality} = \mathbb{E}_{x_k, y_k^* \sim O(x_k)} 
        \mathbbm{1}\left\{ f_{\theta'}(y \mid x_k) = f_{\theta}(y \mid x_k) \right\}
    \end{equation}
    where $O(x_k)$ denotes the locality set for $x_k$, and $f_{\theta}$ is the original model.

    \item \textbf{Fluency} quantifies the entropy of $n$-gram distributions to measure generation diversity and penalize repetition:
    \begin{equation}
        \text{Fluency} = -\frac{2}{3} \sum_{k} g_2(k) \log_2 g_2(k) + \frac{4}{3} \sum_{k} g_3(k) \log_2 g_3(k)
    \end{equation}
    where $g_n(k)$ denotes the frequency of the $k$-th $n$-gram in the generated text.
\end{itemize}

\noindent \textbf{Hyperparameters.}
For the hyperparameter $\alpha$ in our method, we set its value based on the cumulative causal effect of each model module as determined by empirical analysis.
For Qwen2.5-7B, we set $\alpha = 0.3$, with the critical MLP layers and Attn layers selected as $\mathcal{R}_{\mathrm{mlp}} = \{4,\, 5,\, 6,\, 7,\, 8\}$ and $\mathcal{R}_{\mathrm{attn}} = \{2,\, 3,\, 4\}$, respectively. 
For Mistral-7B, we set $\alpha = 0.1$, and choose $\mathcal{R}_{\mathrm{mlp}} = \{4,\, 5,\, 6,\, 7,\, 8\}$ and $\mathcal{R}_{\mathrm{attn}} = \{3,\, 4,\, 5,\, 6\}$. 
These settings are determined according to the causal tracing results and the distribution of factual memory across modules in each model.

\subsection{Batch Knowledge Editing on LLMs}
We systematically conduct batch knowledge editing experiments using \textit{IntAttn-Edit} and baseline methods.
Table~\ref{tab:zsre} summarizes the performance of all methods on the ZsRE benchmark under three different batch sizes: 100, 300, and 500 edits per batch (i.e., $T=100,300,500$).
Table~\ref{tab:counterfact} presents the corresponding results on WikiData counterfact.
\textbf{In all tables, bold values indicate the best performance, and $^\dagger$ marks the second best.}

\begin{table*}[ht]
\centering
\caption{Batch Knowledge Editing Performance on ZsRE.}
\label{tab:zsre}
\resizebox{\textwidth}{!}{
\begin{tabular}{l|cccc|cccc|cccc}
\midrule
& \multicolumn{12}{c}{\textbf{Mistral-7B}} \\
\cmidrule(lr){2-13}
Method
  & \multicolumn{4}{c|}{T=100}
  & \multicolumn{4}{c|}{T=300}
  & \multicolumn{4}{c}{T=500} \\
\cline{2-5} \cline{6-9} \cline{10-13}
  & Edit Succ. & Por. & Loc. & Flu. 
  & Edit Succ. & Por. & Loc. & Flu. 
  & Edit Succ. & Por. & Loc. & Flu. \\
\midrule
IntAttn-Edit & \textbf{92.70} & \textbf{56.25} & 34.88$^\dagger$ & 568.08$^\dagger$ & \textbf{87.75} & \textbf{56.30} & 34.89$^\dagger$ & 564.44$^\dagger$ & \textbf{86.15} & \textbf{55.42} & 34.40$^\dagger$ & 568.20$^\dagger$ \\
FT           & 39.99 & 49.04 & \textbf{45.79} & \textbf{575.94} & 39.99 & 49.04 & \textbf{45.76} & \textbf{575.61} & 39.85 & 48.85  & \textbf{45.75}  & \textbf{575.52}    \\
ROME         &  5.05 &  1.24 &  0.52 & 443.30 &  3.08 &  0.80 &  0.55 & 467.93 &  1.34 &  0.51 &    0.51 & 449.34   \\
R-ROME       &  4.86 &  1.22 &  0.53 & 443.72 &  2.59 &  0.61 &  0.57 & 444.04 &  2.11 &  0.73 &  0.52 &  470.61      \\
MEMIT        & 91.66$^\dagger$ & 55.71$^\dagger$ & 33.88 & 564.61 & 84.29$^\dagger$ & 55.63$^\dagger$ & 33.35 & 563.44 & 83.58$^\dagger$ & 54.22$^\dagger$ & 32.76 & 563.31 \\
AlphaEdit    & 85.10 & 51.44 & 28.76 & 554.43 & 73.16 & 45.50 & 25.83 & 546.26 & 57.49 & 37.83 &    21.06 & 525.05   \\
\midrule
& \multicolumn{12}{c}{\textbf{Qwen2.5-7B}} \\
\cmidrule(lr){2-13}
Method
  & \multicolumn{4}{c|}{T=100}
  & \multicolumn{4}{c|}{T=300}
  & \multicolumn{4}{c}{T=500} \\
\cline{2-5} \cline{6-9} \cline{10-13}
  & Edit Succ. & Por. & Loc. & Flu. 
  & Edit Succ. & Por. & Loc. & Flu. 
  & Edit Succ. & Por. & Loc. & Flu. \\
\midrule
IntAttn-Edit & \textbf{96.98} & \textbf{55.04} & \textbf{34.40} & \textbf{572.78} & 93.96$^\dagger$ & \textbf{55.41} & \textbf{34.59} & 569.67$^\dagger$ & \textbf{92.10} & \textbf{56.32} & \textbf{34.44} & 562.54$^\dagger$ \\
FT           & 25.63 & 26.02 & 17.66 & 243.78 & 28.94 & 33.07 & 25.41 & 276.25 & 30.79 & 34.61 &    26.64& 296.49 \\
ROME         & 79.06 & 39.37 & 19.06 & 561.96 & 25.83 & 8.83  &  3.46 & 484.33 & 19.33 &  4.25 &    1.27&  488.28 \\
R-ROME       & 83.69 & 41.84 & 20.82 & 567.66 & 27.98 & 10.42 &  4.13 & 460.25 & 19.55 &  4.53 &    1.57&  473.55  \\
MEMIT        & 95.43 & 53.88$^\dagger$ & 32.58$^\dagger$ & 569.84$^\dagger$ & 91.44 & 53.94$^\dagger$ & 32.99 & \textbf{570.56} & 91.06$^\dagger$ & 54.74$^\dagger$ & 32.53$^\dagger$ & \textbf{563.33} \\
AlphaEdit    & 96.87$^\dagger$ & 53.62 & 31.71 & 568.88 & \textbf{94.98} & 53.93 & 31.17 & 568.55 & 86.89 & 52.17 & 29.74 & 561.09 \\
\bottomrule
\end{tabular}
}
\end{table*}

\begin{table*}[ht]
\centering
\caption{Batch Knowledge Editing Performance on WikiData counterfact.}
\label{tab:counterfact}
\resizebox{\textwidth}{!}{
\begin{tabular}{l|cccc|cccc|cccc}
\midrule
& \multicolumn{12}{c}{\textbf{Mistral-7B}} \\
\cmidrule(lr){2-13}
Method
  & \multicolumn{4}{c|}{T=100}
  & \multicolumn{4}{c|}{T=300}
  & \multicolumn{4}{c}{T=500} \\
\cline{2-5} \cline{6-9} \cline{10-13}
  & Edit Succ. & Por. & Loc. & Flu. 
  & Edit Succ. & Por. & Loc. & Flu. 
  & Edit Succ. & Por. & Loc. & Flu. \\
\midrule
IntAttn-Edit & \textbf{92.52} & \textbf{55.61} & 36.34$^\dagger$ & 581.60$^\dagger$ & \textbf{84.48} & \textbf{50.15} & 36.23$^\dagger$ & 575.35$^\dagger$ & \textbf{77.81} & \textbf{45.96} & 35.93$^\dagger$ & 570.90$^\dagger$ \\
FT           & 26.03 & 26.63 & \textbf{59.96} & \textbf{594.07} & 26.05 & 26.56 & \textbf{60.07} & \textbf{600.09} & 26.06 & 26.61 &    \textbf{60.00} & \textbf{601.15} \\
ROME         &  4.69 &  2.65 &  1.41 & 491.41 &  0.77 &  0.99 &  0.61 & 556.92 &  1.05 &  1.19 &     0.70 & 556.05 \\
R-ROME       &  4.31 &  2.42 &  1.21 & 513.69 &  0.73 &  1.03 &  0.59 & 552.59 &  0.75 &  1.12 &     0.65 & 544.06 \\
MEMIT        & 90.18$^\dagger$ & 55.44$^\dagger$ & 35.00 & 578.10 & 78.86$^\dagger$ & 47.89$^\dagger$ & 33.43 & 566.03 & 72.16$^\dagger$ & 43.12$^\dagger$ & 32.75 & 559.26 \\
AlphaEdit    & 80.34 & 50.64 & 28.18 & 567.34 & 48.60 & 29.59 & 18.97 & 515.48 & 38.71 & 23.58 &    16.66 & 514.00    \\
\midrule
& \multicolumn{12}{c}{\textbf{Qwen2.5-7B}} \\
\cmidrule(lr){2-13}
Method
  & \multicolumn{4}{c|}{T=100}
  & \multicolumn{4}{c|}{T=300}
  & \multicolumn{4}{c}{T=500} \\
\cline{2-5} \cline{6-9} \cline{10-13}
  & Edit Succ. & Por. & Loc. & Flu. 
  & Edit Succ. & Por. & Loc. & Flu. 
  & Edit Succ. & Por. & Loc. & Flu. \\
\midrule
IntAttn-Edit & 97.92$^\dagger$ & 53.24 & \textbf{33.33} & \textbf{605.23} & \textbf{95.76} & 50.79 & \textbf{33.72} & \textbf{602.80} & \textbf{94.33} & 51.39$^\dagger$ & \textbf{33.81} & \textbf{606.73} \\
FT           & 18.55 & 15.43 & 22.63 & 313.17 & 18.46 & 16.02 & 28.28 & 305.10 & 17.52 & 15.54 &     29.97 & 341.42  \\
ROME         & 32.06 & 14.64 &  7.63 & 486.98 & 12.89 &  3.82 &  1.35 & 498.40 & 16.93 & 4.31  &     1.02 &  522.71      \\
R-ROME       & 40.07 & 18.98 & 11.29 & 501.20 & 14.39 &  4.02 &  1.61 & 512.02 & 18.89 &  4.68 &     1.79 & 558.50 \\
MEMIT        & 97.67 & 54.67$^\dagger$ & 31.80$^\dagger$ & 601.88$^\dagger$ & 95.69$^\dagger$ & \textbf{53.18} & 31.62$^\dagger$ & 600.60$^\dagger$ & 94.19$^\dagger$ & \textbf{53.35} & 32.13$^\dagger$ & 603.20$^\dagger$ \\
AlphaEdit    & \textbf{98.01} & \textbf{56.24} & 31.15 & 598.95 & 94.60 & 53.13$^\dagger$ & 29.44 & 598.50 & 91.67 & 50.12 &    
27.83 & 594.44  \\
\bottomrule
\end{tabular}
}
\end{table*}

Across all experimental settings, \textit{IntAttn-Edit} demonstrates clear advantages in both editing efficacy and generalization. 
On the ZsRE dataset, \textit{IntAttn-Edit} achieves the best Edit Success and Portability scores for every batch size. 
For example, with $T=100$, \textit{IntAttn-Edit} reaches 96.98 Edit Success and 55.04 Portability on Qwen2.5-7B, outperforming the second-best method by 0.11 and 1.42 respectively, while also maintaining high Locality and Fluency. Even as the batch size increases to $T=500$, \textit{IntAttn-Edit} sustains high performance (92.10 Edit Success and 56.32 Portability), while other baselines, such as MEMIT and AlphaEdit, show a more significant decline. 
On Mistral-7B, \textit{IntAttn-Edit} similarly surpasses other methods, maintaining Edit Success above 86 and Portability above 55 across all batch sizes.

The trend is consistent on the WikiData counterfact dataset, where \textit{IntAttn-Edit} continues to lead in most metrics. 
For instance, on Qwen2.5-7B, \textit{IntAttn-Edit} attains 97.92 Edit Success at $T=100$ and outperforms most baselines on Portability and Locality at all edit scales. 
Particularly under large-batch settings (e.g., $T=500$), \textit{IntAttn-Edit} maintains its robustness with only a minor decrease in Edit Success, whereas other methods drop more sharply.
In addition, \textit{IntAttn-Edit} balances the tradeoff between Edit Success and knowledge retention, with Locality and Fluency scores comparable to or surpassing baselines, indicating minimal side effects on unrelated knowledge and output fluency.

\subsection{Analysis of the Knowledge Balancing Strategy}

We further analyze the effectiveness of the proposed knowledge balancing strategy in \textit{IntAttn-Edit}, which dynamically determines the relative contributions of Attn and MLP modules during knowledge editing. To assess its impact, we conduct systematic experiments on the ZsRE dataset using advanced LLMs, varying the balancing weights between modules and evaluating performance across four standard metrics: Edit Success, Portability, Locality, and Fluency. 
For completeness, as clarified in Section~5.3, we also evaluate the two endpoints of the balance factor: $\alpha=1$ applies updates solely to the attention modules (\textit{Attn-only}), whereas $\alpha=0$ applies updates solely to the MLP modules (\textit{MLP-only}); the full endpoint results are provided in Appendix~B.2 of the supplementary material.

\begin{figure}[htbp]
\centering
\includegraphics[width=0.8\textwidth]{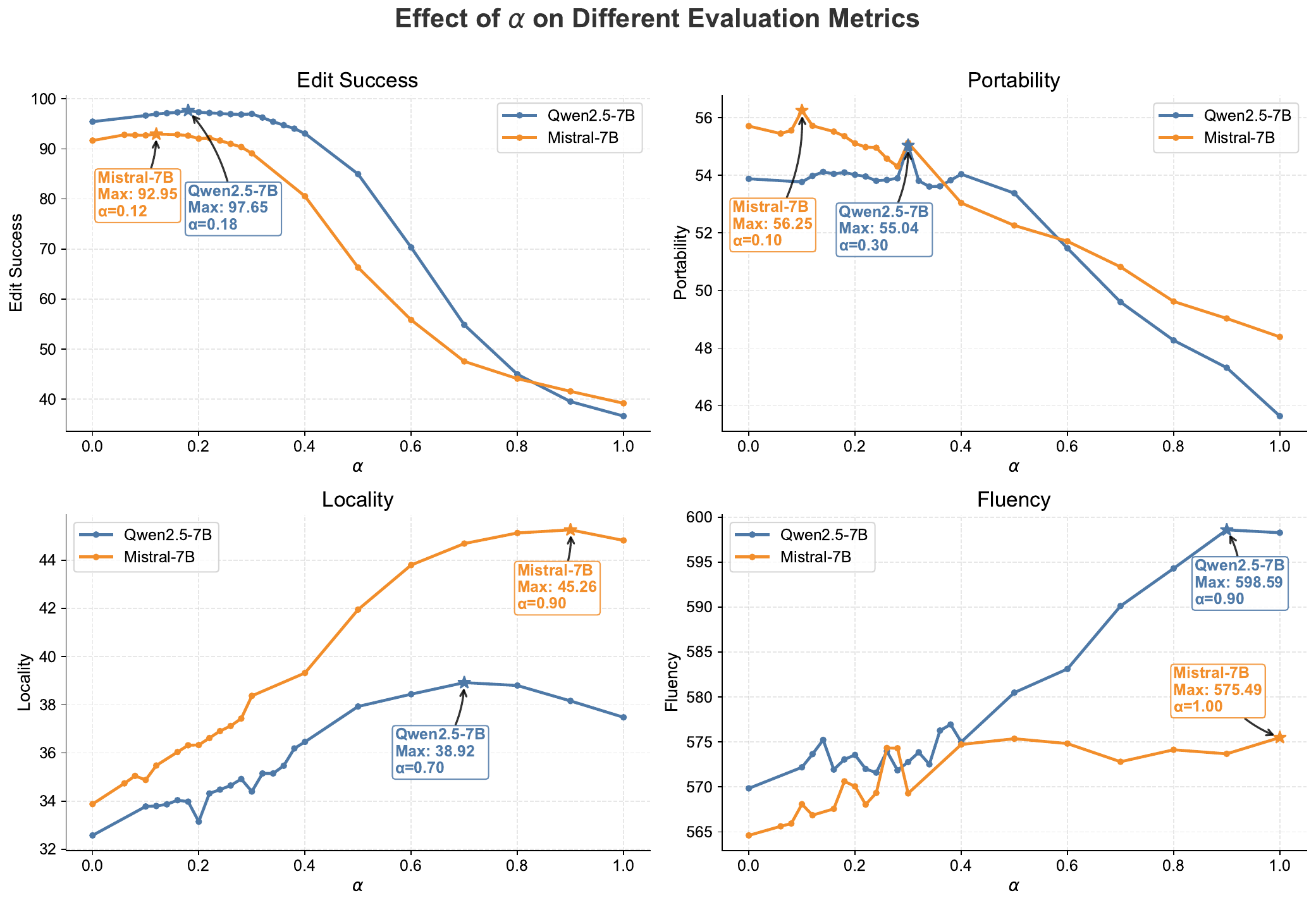}
\caption{Effect of the knowledge balancing strategy on evaluation metrics using Qwen2.5-7B and Mistral-7B models on the ZsRE dataset. Optimal values are highlighted for clarity.}
\label{fig:alpha_metrics_compare}
\end{figure}

As shown in Figure~\ref{fig:alpha_metrics_compare}, assigning moderate weights to the Attn module achieves the best trade-off between factual accuracy, generalization, and knowledge preservation. Both models reach peak Edit Success and Portability at moderate balancing ratios, demonstrating that our causal-effect-guided allocation effectively harmonizes the strengths of both modules. Although increasing the relative weight of Attn can lead to marginal improvements in Locality and Fluency, these gains are offset by declines in Edit Success and Portability. Overall, these results confirm that our knowledge balancing strategy maintains model editing performance within the optimal range, enabling robust, reliable, and generalizable knowledge updates in advanced LLMs.

\section{Conclusion}

In this work, we address a fundamental limitation of existing knowledge-editing approaches, namely the issue of knowledge residuals caused by updating only a subset of knowledge-storage modules. 
Unlike previous methods that focus solely on the MLP module.
We propose for the first time to jointly edit MLP and Attn modules in large language models. 
Our comprehensive causal tracing experiments reveal that Attn modules contribute substantially to factual knowledge storage and retrieval, in addition to the well-recognized role of MLP modules. 
Building on these insights, we introduce \textit{IntAttn-Edit}, an innovative method that extends the associative memory-based editing paradigm to Attn modules. 
By dynamically allocating update magnitudes between MLP and Attn modules according to their measured causal contributions, \textit{IntAttn-Edit} achieves more balanced and comprehensive knowledge updates. 
Extensive experiments on benchmarks such as ZsRE and WikiData Counterfact demonstrate that our approach consistently outperforms existing methods, ensuring higher editing efficacy and stronger knowledge preservation across various editing scenarios.


\bibliography{acml25}
\let\clearpage\relax

\end{document}